\documentclass{article}
\usepackage[numbers]{natbib}
\usepackage {arxiv}

\usepackage{cite}
\usepackage{floatrow}
\usepackage{amsmath,amssymb,amsfonts}
\usepackage{algorithm}
\usepackage{algorithmic}
\usepackage{textcomp}
\usepackage{xcolor}
\usepackage{float}
\usepackage{booktabs}
\usepackage{multirow}
\usepackage{enumitem}
\usepackage{setspace}
\usepackage[]{graphicx}
\usepackage{array}
\usepackage{amsmath}
\usepackage{amssymb}
\usepackage{epsfig}
\usepackage{epstopdf}
\usepackage{subfigure}
\usepackage{comment}
\usepackage{tabularx}
\usepackage{threeparttable}
\usepackage{adjustbox}
\usepackage{soul}
\usepackage{dsfont}
\usepackage[hyphens]{url}
\usepackage{hyperref} 

\title{\LARGE \bf
LATTLE: LLM Attention Transplant for Transfer Learning of Tabular Data Across Disparate Domains}

\author{Ibna Kowsar, Kazi F. Akhter and Manar D. Samad\\
Department of Computer Science\\
Tennessee State University\\
Nashville, TN, USA\\
\texttt{msamad@tnstate.edu} \\
 }
    
\begin{document}

\maketitle

\begin{abstract}

Transfer learning on tabular data is challenging due to disparate feature spaces across domains, in contrast to the homogeneous structures of image and text. Large language models (LLMs) offer a knowledge base to improve the limited effectiveness of cross-domain transfer learning for tabular data. However, LLM performance often stagnates due to subjective text prompts and the computational limitations of in-context learning. We present a novel language-to-tabular context-learning method that uses attention-specific transformer weights, enabling seamless transfer learning across disparate tabular data sets. The LLM attention transplant mechanism facilitates a domain-agnostic transfer learning, eliminating the need for shared features between tables, LLM prompt engineering, and large-scale pretrained models. Our experiments using ten pairs of disjoint source-target data sets and 12 baseline methods demonstrate the superiority of the proposed LLM-attention transplant for transfer learning (LATTLE) method over traditional ML models, state-of-the-art deep tabular architectures, and models trained on thousands to billions of tabular samples. The proposed cross-domain attention transfer demonstrates an effective solution for adapting LLMs to learning non-text tabular data in a low-resource environment. The source code of the proposed method is shared in
\footnote{\href{https://anonymous.4open.science/r/LATTLE---LLM-Tabular-Transfer-Learning-667E}%
{https://anonymous.4open.science/r/LATTLE---LLM-Tabular-Transfer-\allowbreak Learning-667E}}

\end{abstract}


\section{Introduction}

Tabular data sets are structured in rows of samples and columns of mixed-type features, which are ubiquitous in spreadsheets and relational databases in countless domains. Despite widespread applications, tabular data have not attracted much attention in artificial intelligence (AI) research compared to image and text data. The recent AI revolution has ushered in a new era through deep learning (DL) of image and text, introducing large language models (LLMs), vision transformers (ViTs), and vision language models (VLMs). In contrast, a survey on deep learning of tabular data~\citep{Borisov2022_Survey} reports that tabular data still rely on traditional machine learning (ML), particularly XGBoost~\citep{chen2016xgboost} and LightGBM~\citep{ke2017lightgbm}, due to their superior performance over deep learning methods. Similar findings dissuade practitioners from reaping the benefits of deep representation and transfer learning for tabular data sets. 

Deep learning of tabular data is challenging for three reasons. First, the inductive bias required for robust representation learning of tabular data with mixed data types is not as developed as that of image and text data. Tabular data learning with mixed data types still relies on feature engineering with traditional ML. Second, many application domains have tabular data with limited samples for deep learning, whereas tabular data augmentation is not a trivial task~\citep{kowsar2023buet_deepCluster, rabbani2024_gceals}. Third, learning the relationship between tabular data sets of different domains is nontrivial due to the lack of transferable patterns or context in deep representations, similar to image and text data. Therefore, deep representation learning of shared knowledge and transferable context can pave the way for robust transfer learning of tabular data sets.

The recent revolution in AI has demonstrated the versatility of large language models (LLMs) beyond text generation, even with minimal or no training samples. LLMs, as large reservoirs of knowledge, can perform zero- and few-shot learning of tabular data ~\citep{Han2024_featLLM, Huertas2024ImproveLGMonFewShots, Hegselmann2023_TabLLM}. However, zero- and few-shot learning approaches perform poorly compared to traditional ML~\citep{Abdul2024_TabLM}. An efficient transfer learning method for tabular data can take full advantage of the knowledge base of LLM foundation models. This paper presents a novel cross-attention learning framework using LLMs that enables transfer learning between tabular data of different domains. The cross-attention between an LLM fine-tuned on the source data and a gated Feature Tokenizer Transformer (gFTT) is established by transplanting the frozen LLM weights into the gFTT for downstream classification of target data.  

The organization of the paper is as follows. Section~\ref{subsec: related_work} presents the latest deep learning and LLM approaches proposed in the context of tabular data and transfer learning. Section~\ref{sec: method} discusses the preliminaries on attention-based computing and the proposed LLM attention transplant framework for transfer learning. Section~\ref{sec: experiments} discusses the tabular data sets, the experimental setup and scenarios, and the evaluation method. Section~\ref{sec: results} presents the transfer learning performance of the proposed method and compares it with other leading baselines for learning tabular data. Section~\ref{sec: Discussion} summarizes the key findings, provides insight into the results, and outlines the limitations. The paper concludes in Section~\ref{sec: Conclusions}.
  
\subsection {Literature review and contributions} \label{subsec: related_work}
Traditional machine learning (ML) methods still have strong dominance in tabular data application domains due to their effectiveness in handling features of mixed data types~\citep{Grinsztajn2022}. Gradient-boosted decision trees, such as XGBoost~\citep{chen2016xgboost}, perform well for structured tabular data, particularly when the data set is small. Recent deep learning methods, including TabNet~\citep{Arik2021_TabNet}, Feature Tokenizer Transformer (FTT)~\citep{Gorishniy2021_FTT}, and Gated Feature Tokenizer Transformer (gFTT)~\citep{Wang2022_Transtab}, have been developed specifically to learn from tabular data. These methods demonstrate the promising performance of self-attention mechanisms, outperforming traditional ML methods in numerous instances with large sample sizes. TabPFN~\citep{hollmann2022_TabPFN} proposes a tabular foundation model for small to medium-sized data sets using a probabilistic neural network approach. However, the model requires a vast amount of synthetic data during pretraining. 

A large pretrained model is often used in transfer learning when the sample size is too small for effective deep learning. In transfer learning, a model is first pretrained using a source data set with a large sample size, which is then fine-tuned using a target data set with limited samples. Several transfer learning methods have recently been proposed for tabular data sets. TransTab~\citep{Wang2022_Transtab} is a state-of-the-art method that requires common features between the source and target data sets to align the data representations for knowledge transfer. In practice, two tabular data sets from different domains rarely share common features. Levin et al.~\citep{levin2023_Tabular-transfer} propose a similar transfer learning method between tabular data sets of the same domain with shared features. Ye et al. have introduced the cross-table mask modeling (CM2) framework~\citep{Ye2024_CM2} to learn generalizable data representations in heterogeneous feature spaces and tabular data domains, thus enhancing adaptation during fine-tuning. Similar to the CM2 method, XTab~\citep{Zhu2023_Xtab} pretrains an FTT model using tabular data sets from diverse domains. However, their downstream classification through fine-tuning often performs poorly compared to decision tree models (e.g., XGBoost~\citep{chen2016xgboost}). Therefore, existing methods struggle to optimally perform transfer learning for tabular data, particularly when the source and target schemas differ significantly. Furthermore, the realistic but challenging transfer learning between tabular data sets without common features has not been well studied. 

\begin{figure*}[t]
\centerline{\includegraphics[width=1.0\textwidth]{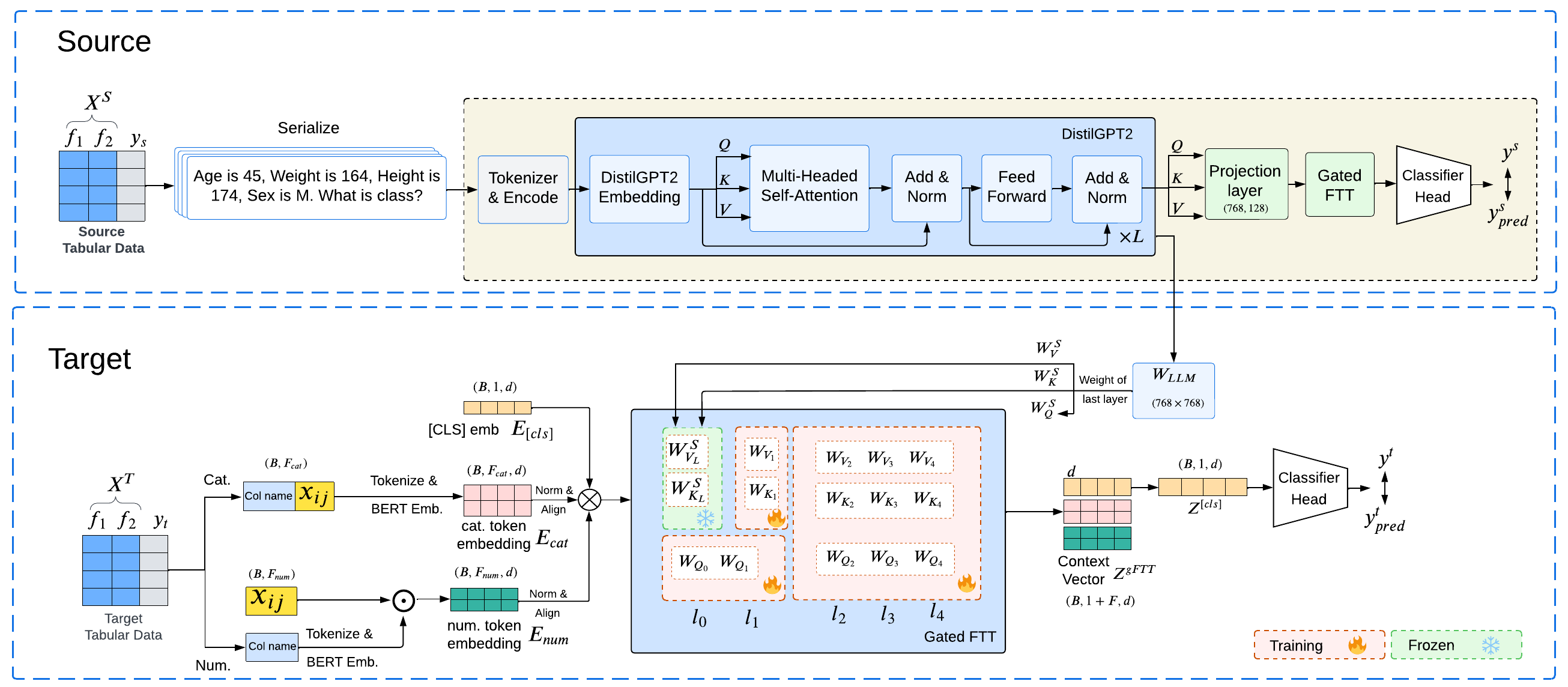}}
\vspace{10pt}
\caption{The proposed LLM attention transplant for transfer learning (LATTLE) framework. The attention-related weights are transplanted from a lightweight LLM (DistilGPT2) fine-tuned by source data to a Gated Feature Tokenizer Transformer (gFTT) to be fine-tuned using target data. Emb. = Embedding, Cat. = categorical and Num. = numerical features.  
}
\label{fig:lattle_arch}
\vspace{-10pt}
\end{figure*} 

In this context, recent breakthroughs in large language models (LLMs) offer a new paradigm of learning and continue to produce unprecedented examples of data-driven knowledge discovery. Zeng et al. demonstrate that LLM embeddings can handle distribution shifts in data sets belonging to the same domain~\citep{zeng2025llm-same-domain-TL}. However, effective integration of non-text data into text-based LLMs remains an open problem. A recent benchmark study on tabular language models (TabLM) reveals that traditional ML frequently outperforms LLMs in tabular data, possibly due to the tabular-to-text serialization required by text-based LLMs~\citep{Abdul2024_TabLM}. Notably, tabular-to-text data serialization is a critical step in LLM prompt engineering~\citep{brown2020_language-models-fews-hot-learners, Han2024_featLLM}. Most applications of LLM in tabular data rely on in-context learning (ICL), where LLMs perform tasks by observing a few serialized labeled samples with text instructions, without updating the model parameters~\citep{shi2025_LATTE, Han2024_featLLM, nam2024_P2T, Wang2023a_MediTab}.

Nam et al.~\citep{nam2024_P2T} propose Prompt to Transfer (P2T) to conduct LLM-based transfer learning between data tables with common features. They identify the feature in the unlabeled source data that has the highest correlation with the target data label. The ICL prompt uses the correlated feature as a pseudo-label for source data examples. In contrast, target data examples in the prompt include the correlated feature of the source to predict the target label. We identify three limitations of the ICL approach to LLM. First, ICL does not update the model weights necessary for effective knowledge transfer. Second, ICL, as used in the P2T method, requires tedious and manual prompt engineering for each source–target pair. Third, the limit on token size in the prompt restricts the number of data examples in ICL. Studies have also shown that LLM performance often degrades when the number of few-shot examples exceeds 32~\citep{chen2022_limitation-ICL, Gardner2024_TABULA-8B, spinaci2025ConTextTab}. Therefore, optimizing the LLM weights for data-specific tasks can overcome the limitations of ICL.

Recent studies have optimized LLM weights by fine-tuning~\citep{Yang2024_iTabLLM, Dinh2022_LIFT}. Gardner et al.~\citep{Gardner2024_TABULA-8B} propose TABULA-8B, a LLaMA-3-based language model fine-tuned on billions of tabular data rows serialized as text. LLM fine-tuning in similar studies uses an Auto-Regressive loss that predicts the next token in the sequence~\citep{Gardner2024_TABULA-8B, Dinh2022_LIFT, Zhang2023_LLaMAGTL, Hegselmann2023_TabLLM}, which differs from the learning objective of a classifier model. Furthermore, the pretrained TABULA-8B model supports few-shot in-context learning on downstream target data, but performance gains plateau after 32 shots in the prompt due to the inherent limitations of ICL. We argue that downstream classification can also involve model fine-tuning to overcome the limitations of ICL. Moreover, their pretrained model is trained on eight billion tabular data rows, requiring extensive computing resources, whereas a lightweight transfer learning approach can achieve better performance.

\subsection{Contributions}
In this paper, we propose a new LLM Attention Transplant for Transfer LEarning (LATTLE) to address the challenges of cross-domain tabular data transfer learning using cross-attention with an LLM. The contributions of our paper are as follows. 

\begin{itemize}
    \item We propose one of the first cross-attention transfer learning methods between an LLM and a feature tokenized transformer for tabular data.
    \item The LLM is trained using a learning objective designed for tabular data, rather than the standard auto-regressive loss that predicts the next token in a sequence. 
    \item The context between heterogeneous domains is captured by transplanting attention weights from an LLM, which obviates the requirement for shared features between different tabular datasets.
    \item The proposed method performs downstream cross-attention learning using finetuned LLM weights without the need for prompt engineering and in-context learning.
     \item The efficacy of the proposed attention transplant is demonstrated using a lightweight LLM and a decently sized source data set without requiring millions of samples and heavy computing resources.
\end{itemize}

\section {Methodology} \label{sec: method}

This section provides essential background on the attention mechanism and the proposed attention transplant approach to achieve LLM-based transfer learning of tabular data. 


\subsection{Preliminaries} 

A tabular data set is mathematically represented as a matrix \( X \in \mathbb{R}^{N \times m} \) with \( N \) rows of samples and \( m \) columns of feature variables characterizing individual samples. Each matrix element \( X_{ij} \) represents the value of the \( j \)-th feature of the \( i \)-th sample. Attention between features, namely self-attention, needs to convert each feature token into a vector representation, known as an embedding. A BERT tokenizer converts each input token into an embedding of dimension \( d \). Let \( f_j \in \mathbb{R}^{d \times 1} \) denote the embedding of the \( j \)-th feature of a sample. A transformer is trained to obtain the three vectors related to attention ($query$, $key$, and $value$) from the input embedding, as shown in Equation~\ref{eq:qkv_featurewise}.
\begin{equation}
q_j = W_qf_j , \quad k_j = W_kf_j , \quad v_j = W_vf_j.
\label{eq:qkv_featurewise}
\end{equation}
Here, \{$W_q$, $W_k$, $W_v$\} $\in \mathbb{R}^{d_k \times d}$ are trainable weight matrices, and \{$q_j$, $k_j$, $v_j$\} $\in \mathbb{R}^{d_k}$ are the resulting $query$, $key$, and $value$ vectors for feature $j$. The attention of feature $p$ to feature $j$ ($\alpha_{pj}$) is computed using a scaled dot product of the $query$ vector of the $p$ feature and the $key$ vector of the $j$ feature. 
\begin{equation}
\alpha_{pj} = \text{softmax} \left( \frac{q_p^\top k_j}{\sqrt{d_k}} \right), \quad 
Z_p = \sum_{j=1}^{m} \alpha_{pj} v_j
\label{eq:weighted_sum}
\end{equation}
Here, \( Z_p \in \mathbb{R}^{d_k} \) represents the attention-weighted context vector of feature $p$ across all $m$ features. It should be noted that attention weights are directional because \( \alpha_{pj} \neq \alpha_{jp} \).

\begin{table*}[t]
\centering
\caption{Summary of source and target data sets used to evaluate cross-domain transfer learning.}
\label{tab:data_sets}
\scalebox{0.7}{
\begin{tabular}{llllccccc}
\toprule
Type & Data set & Abbr. & Domain & Samples & Features & Numeric & Categorical & Classes \\
\midrule
\multirow{9}{*}{\shortstack{Source\\Data sets}} 
& mfeat-fourier           & MF   & Shape Measurements & 2000  & 76 & 76 & 0  & 10 \\
& credit-g                & CG    & Banking            & 1000  & 20 & 7  & 13 & 2  \\
& sick                    & SK  & Thyroid Disease    & 3772  & 29 & 7  & 22 & 2  \\
& optdigits               & DG  & Optical Digits   & 5620  & 64 & 64 & 0  & 2  \\
& steel-plates-fault      & SP   & Manufacturing      & 1941  & 33 & 33 & 0  & 2  \\
& car-evaluation          & CE   & Car Pricing        & 1728  & 21 & 0  & 21 & 4  \\
& churn                   & CH  & Telecommunication  & 5000  & 20 & 16 & 4  & 2  \\
& cardiovascular-disease  & CD   & Heart Disease      & 70000 & 11 & 5  & 6  & 2  \\
& seismic-bumps           & SB  & Hazard Monitoring  & 2584  & 18 & 14 & 4  & 2  \\
\midrule
\multirow{5}{*}{\shortstack{Target\\Data sets}} 
& cmc                     & CM   & Demographics       & 1473  & 9  & 2  & 7  & 3  \\
& diabetes                & DB  & Metabolic Disease  & 768   & 8  & 8  & 0  & 2  \\
& vehicle                 & VH   & Automotive         & 846   & 18 & 18 & 0  & 4  \\
& pc1                     & PC1   & Software Testing   & 1109  & 21 & 21 & 0  & 2  \\
& cylinder-bands          & CB    & Industrial Design  & 540   & 39 & 18 & 21 & 2  \\
\bottomrule
\end{tabular}}
\vspace{-10pt}
\end{table*}

\subsection{Fine-tuning LLM using source tabular data} \label{llm-finetune}

The first step of the proposed transfer learning method is to adapt an LLM to a tabular data learning task. In particular, LLMs are trained using an autoregressive loss to predict the next token in a sentence. Adapting language models to mixed-type tabular data requires a tailored data preprocessing framework. The first step involves serialization of feature value pairs of tabular data in a textual description (e.g. ``Age is 25. Sex is male.''), which are then tokenized and encoded, as shown in Figure~\ref{fig:lattle_arch}. It is important to note that the seminal GPT‑3.5 model and its successor LLMs are proprietary and not open-source. Although more advanced open-source LLMs can be finetuned with expensive computational resources, they may have already been trained on publicly available data, making them prone to data memorization. Accordingly, we adopt DistilGPT2~\citep{huggingface_distilgpt2}, an open-source, lightweight variant of GPT2~\citep{radford2019_GPT2}, to demonstrate the effectiveness of language-to-tabular cross-attention learning. DistilGPT2 reduces the computational cost by 50\% while retaining approximately 97\% of the original performance of GPT2. DistilGPT2 has six attention layers and accepts input sequences of up to 1024 tokens. The tokenizer (Byte Pair Encoding (BPE)) of DistilGPT2 is used to obtain token IDs and the corresponding embedding vectors from pretrained DistilGPT2. To enable LLM-to-tabular context learning, we adopt a state-of-the-art transformer designed for tabular data, the gated feature tokenizer transformer (gFTT) ~\citep{Wang2022_Transtab} together with DistilGPT2, forming an end-to-end learning framework. As transformer-based architectures, both gFTT and DistillGPT2 possess trainable projection matrices associated with the $key$, $query$, and $value$ representations (see Equation~\ref{eq:qkv_featurewise}). These projection weights are used to streamline language-to-tabular knowledge transfer within the DistilGPT2-gFTT joint framework, which is followed by a classifier head. The top layer of DistilGPT2 is fine-tuned jointly with gFTT using the source tabular data to perform a supervised classification task defined by the cross-entropy loss in Equation~\ref{eq:ce}.
\begin{equation}\label{eq:ce}
\mathcal{L}_{\text{CE}} = - \frac{1}{n} \sum_{i=1}^{n} \sum_{c=1}^{C} y_{ic} \log \hat{p}_{ic}
\end{equation}
Here, $n$ denotes the number of training samples, $C$ is the number of classes, $y_{ic} \in \{0,1\}$ indicates the ground truth label, and $\hat{p}_{ic}$ denotes the predicted probability for class $c$ given input $i$.

\begin{algorithm}[t]
\renewcommand{\footnotesize}{\normalsize}
\caption{The LATTLE Algorithm}
\label{alg:lattle}
\begin{algorithmic}
\STATE \textbf{Models:} LLM (Source), gFTT (Head), gFTT (Target)
\STATE \textbf{Input:} Source data ($X_s$, $y_s$), Target data ($X_t$, $y_t$)
\STATE \textbf{Output:} $gFTT_{CA}$ cross-table transfer via LLM attention
\vspace{0.5em}
\STATE \textbf{LLM Fine-tuning on Source: Language-to-Tabular Alignment}
\STATE LLM-gFTT $\leftarrow$  \{LLM (Source), gFTT (Head)\}
\FOR{$\text{epoch} = 1$ $\rightarrow$ $n_{\text{epochs}}$}
    \STATE $\mathcal{L}^{Source}_{\mathrm{CE}} \leftarrow \mathcal{L}(W^{\mathrm{LLM}}_{r,L}, \{W^{gFTT}_{r,\ell}\}; X_s, y_s)$ (Eq.~\ref{eq: loss_update_source})    
    \STATE Update LLM and gFTT Attention Weights $\leftarrow$ $W^{LLM}_{r,L}$, $W^{gFTT}_{r,\ell}$ (Eq.~\ref{eq: llm_update},~\ref{eq: fgtt_source_update)})
\ENDFOR
\vspace{0.5em}
\STATE Weights of $L$ Attention Layers of $LLM$ : $\{W^1, ..., W^L\}$ 
\vspace{0.5em}
\STATE \emph{Key-Value} Projection Weights from LLM: $\{W^{LLM}_{K,L}, W^{LLM}_{V,L}\}$ $\leftarrow$ $W^{LLM}_{r,L}$
\vspace{0.5em}
\STATE \textbf{gFTT (Target) as $t$ : Source-to-Target Transfer Learning}
\vspace{0.5em}
\vspace{0.5em}
\STATE \quad $W^t_{K,0} \leftarrow W^{LLM}_{K,L}; \quad W^t_{V,0} \leftarrow W^{LLM}_{V,L}$ \textit{(Frozen)} 
\vspace{0.5em}
\FOR{$\text{epoch} = 1$ $\rightarrow$ $n_{\text{epochs}}$}
    \STATE Cross-attention, $\alpha_{s, t} \leftarrow \text{Attention}(W_q^t, W_k^{LLM}, f_t)$ (Eq.~\ref{eq:cross-attention})   
    \STATE Attention weighted embedding, $Z_t \leftarrow \alpha_{s,t} W_v^{LLM} f_t$ (Eq.~\ref{eq:attentionweighted})    
    \STATE $\mathcal{L}^{Target}_{\mathrm{CE}} \leftarrow \ell(W^t_{Q,0}, \{W^t_{r,\ell}\}_{\ell=1}^4; X_t, y_t)$ (Eq.~\ref{eq: loss_update_target})
    \STATE Update $\{W^{t}_{Q,0}, W^{t}_{r,\ell>0}\}$ minimizing $\mathcal{L}^{Target}_{\mathrm{CE}}$ (Eq.~\ref{eq: zero_gradients})
\ENDFOR
\end{algorithmic}
\end{algorithm}

\subsection{Hierarchical attention weight selection}
Hierarchical representation learning has been well studied in convolutional neural networks for vision tasks. However, the distribution of knowledge in the attention layers of transformers for tabular data remains underexplored. We use empirical evidence on knowledge distribution from the literature to select and transfer attention weights for computing cross-domain (LLM-to-tabular) attention learning. The LLM trained using source data has finetuned projection weights $W_Q^S$, $W_K^S$, and $W_V^S$ corresponding to the \emph{query}, \emph{key}, and \emph{value} representations, respectively. Among these, the \emph{key} and \emph{value} projection weights ($W_K^S$, $W_V^S$) encode contextual representations learned from the source tabular data (Equation~\ref{eq:qkv_featurewise}). The literature suggests that the representations learned by a transformer are distributed hierarchically across layers, where higher layers tend to capture richer contextual knowledge~\citep{dai2022}. Therefore, we use the \emph{key} and \emph{value} projection weights ($W_K^S$, $W_V^S$) of the topmost attention layer of LLM for downstream attention transplantation. Similarly, the downstream transformer model learns target tabular data, where the most contextual target knowledge is at the topmost layer. In contrast, the base, or lowest, layer encodes more general knowledge and is typically kept frozen to preserve this information during downstream transfer learning. This approach is empirically supported by~\citep{hwang2025_freezinglower}, which shows that freezing the bottom layers enhances performance in downstream training. Consequently,  the base layer of the downstream transformer model remains frozen with the LLM weights ($W_K^S$, $W_V^S$), allowing the upper layers to incorporate generalized LLM context while also learning task-specific knowledge for the target tabular data. The LLM-to-tabular context learning mechanisms are explained in the next section.

\subsection{Proposed LATTLE algorithm}

The LLM-based transfer learning method proposed for tabular data is summarized in Algorithm~\ref{alg:lattle} and illustrated in Figure~\ref{fig:lattle_arch}. We learn the language-to-tabular context by finetuning a joint LLM-gFTT framework using source tabular data, as mentioned in Section~\ref{llm-finetune}. The transformer projection weights ($W_q$,$W_k$,$W_v$) of LLM (topmost layer, $L$) and gFTT (all five layers, $l$) are updated to optimize cross-entropy loss as follows.

\begin{equation} \label{eq: llm_update}
W^{LLM}_{r,L}
\leftarrow
W^{LLM}_{r,L}
-
\eta
\frac{\partial \mathcal{L}_{\mathrm{CE}}(X_s)}
{\partial W^{LLM}_{r,L}} ;\quad \forall_r \in \{Q,K,V\}
\end{equation}

\begin{equation} \label{eq: fgtt_source_update)}
W^{gFTT}_{r,\ell}
\;\leftarrow\;
W^{gFTT}_{r,\ell}
\;-\;
\eta
\frac{\partial \mathcal{L}_{\mathrm{CE}}(X_s)}
{\partial W^{gFTT}_{r,\ell}};\quad 
\forall_\ell \in \{0,\dots,4\},\ \forall_r \in \{Q,K,V\}
\end{equation}

The projection weights of the transformers in LLM and gFTT are jointly updated to align the LLM with tabular data representations of gFTT optimizing a common cross-entropy loss in Equation~\ref{eq: loss_update_source} using source data ($X_s$). 

\begin{equation} \label{eq: loss_update_source}
\mathcal{L}^{Source}_{\mathrm{CE}}
=
\mathcal{L}_{\mathrm{CE}}
\Big[
y_s, f (W^{\mathrm{LLM}}_{r,L},\;
\{W^{gFTT}_{r,\ell}\}_{\ell=0}^{4}, X_s)\Big], \forall_r \in \{Q,K,V\}
\end{equation}

The training prepares the projection weights of the LLM for learning the general context of tabular data. The finetuned LLM weights corresponding to $key$ and $value$ representations ($W_k$,$W_v$) contain a general knowledge base suitable for learning tabular data. Therefore, this weight pair is kept frozen at the base of a new downstream gFTT to provide the foundational context of the LLM while updating the upper layers using the target tabular data. In the base layer of the gFTT transformer, the frozen projection weights ($W_k^{LLM}$,$W_v^{LLM}$) of the LLM interact with the unfrozen weight of the $query$ representation ($W_q^t$). Therefore, cross-domain attention $\alpha_{s, t}$ between the LLM tuned to the source and the target is achieved at the weight level of the transformer, as shown in Equation~\ref{eq:cross-attention}, instead of using $key$, $value$, $query$ representations in Equation~\ref{eq:weighted_sum}. 

\begin{equation}
\alpha_{s, t} = \text{softmax} \left( \frac{(W_q^t f_t)^\top W_k^{LLM}f_t}{\sqrt{d_k}} \right). \quad 
\label{eq:cross-attention}
\end{equation}
Here, $f_t$ is the embedding of the target tabular data, explained in the next section. In other words, the target $query$ to LLM $key$ achieves language-to-tabular attention, which is used to achieve cross-domain tabular transfer learning. Attention weighted ($\alpha_{s, t}$) target data embedding ($Z_t$) is obtained following Equation~\ref{eq:attentionweighted}.
\begin{equation}
Z_t = \alpha_{s,t} W_v^{LLM}f_t
\label{eq:attentionweighted}
\end{equation}

The downstream gFTT is optimized for the classification task specific to the target tabular data set. 

\begin{equation} \label{eq: loss_update_target}
\mathcal{L}^{Target}_{\mathrm{CE}}
=
\mathcal{L}_{\mathrm{CE}}
\Big[
y_t, f (W^t_{Q,0},\;
\{W^{t}_{r,\ell}\}_{\ell=1}^{4}, X_t)\Big], \forall_r \in \{Q,K,V\}, 
\end{equation}
where, LLM weight transplantation, $W^t_{K, 0}$ = $W^{LLM}_{K, L}$ and   $W^t_{V, 0}$ = $W^{LLM}_{V, L}$, is performed in downstream with the constraint shown in Equation~\ref{eq: zero_gradients}.
\begin{equation} \label{eq: zero_gradients}
\frac{\partial \mathcal{L}_{\mathrm{CE}}^{Target}}
{\partial W^{t}_{K,0}} =0, 
\frac{\partial \mathcal{L}_{\mathrm{CE}}^{Target}}
{\partial W^{t}_{V,0}}
=
0
\end{equation}

\section {Language-encoded tabular embedding}

The gFTT model in Figure~\ref{fig:lattle_arch} receives embeddings of numerical and categorical values of the target tabular data as inputs. For categorical features, the feature name and value (e.g. ``Sex male'') are tokenized to obtain the corresponding $d$-dimensional embeddings using a pretrained BERT model. The embeddings of $n$ tokens are concatenated to form a categorical feature representation \( E_{\text{cat}} \). For numerical features, the $d$-dimensional BERT embedding of the feature name is multiplied by the numerical value to represent the embedding \( E_{\text{num}} \). A special classification token \([CLS]\) is initialized as a $d$-dimensional embedding \( E_{[\text{cls}]} \). All embeddings \( E_{\text{cat}} \) $\in \mathbb{R}^{B \times F_{cat} \times d}$, \( E_{\text{num}} \) $\in \mathbb{R}^{B \times F_{num} \times d}$ and \( E_{[\text{cls}]} \) $\in \mathbb{R}^{B \times 1 \times d}$ for a batch of B samples are concatenated to represent the input embedding \( E \)$\in \mathbb{R}^{B \times (F+1) \times d}$ for gFTT.

The gFTT model with frozen transplant weights is trained using target data while updating the corresponding projection weight \( W_Q^T \) for $query$ in the lowest attention layer. Following Equation~\ref{eq:weighted_sum}, the target data set is used to query its attention to the source pre-trained LLM through the frozen projection weights (\( W^{S}_K\) and \( W^{S}_V\)). Except for the lowest layer, all gFTT weights (\( W_Q, W_K, W_V \)) are fine-tuned to adapt to the target data distribution while preserving structural priors from the source domain. The gFTT transformers produce context vectors for each feature in the tensor \( Z \)$\in \mathbb{R}^{B \times (F+1) \times d}$. The context vector corresponding to the CLS token \( Z^{[\text{CLS}]}\)$\in \mathbb{R}^{B \times 1 \times d}$ is used as input to the classifier head. In the process of attention computing, 
\( Z^{[\text{CLS}]} \) aggregates the context vectors of all numerical and categorical features into a d-dimensional embedding to represent a sample. A linear classifier head converts \( Z^{[\text{CLS}]} \) into a logit vector over the target classes. The final output probabilities are obtained by applying a softmax function over the logits. The model is trained using cross-entropy loss (Equation~\ref{eq:ce}), applied here to the logits predicted from \( Z^{[\text{CLS}]} \).

\begin{figure*}[t]
\centering
\subfigure[LLM finetuning on source (Credit-g data set)] {
    \includegraphics[width=0.48\textwidth, height=5cm]{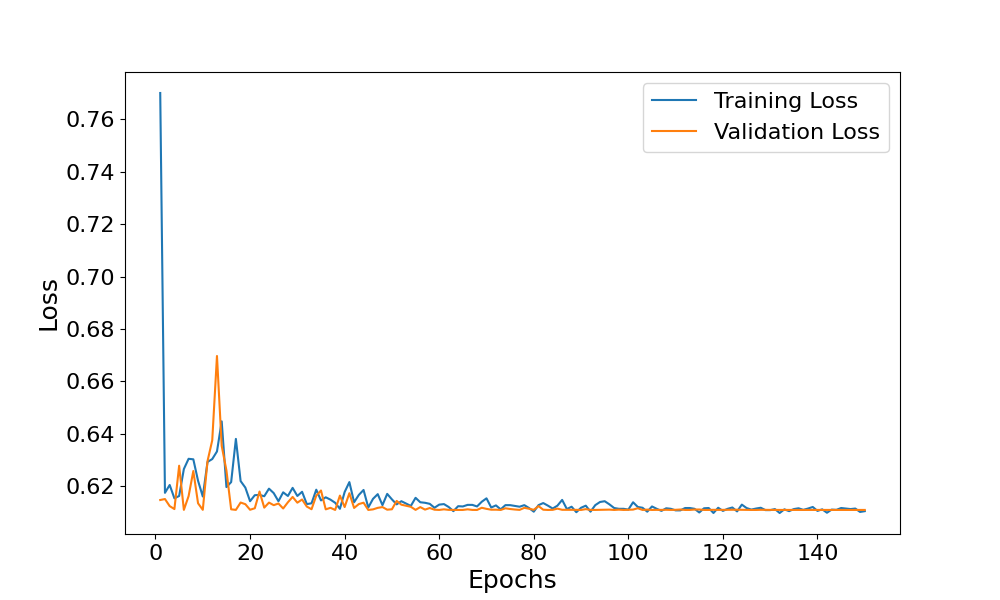}
    \label{fig:sup-pretrain}
}
\hspace{-10pt}
\subfigure[Transfer Learning gFTT on target (Diabetes data set)] {
    \includegraphics[width=0.48\textwidth, height=5cm]{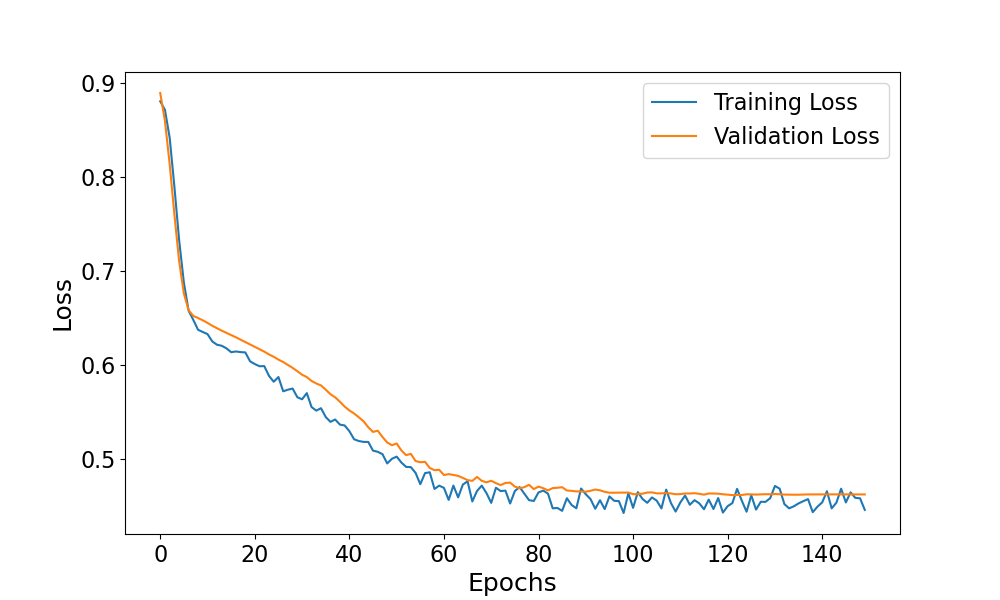}
    \label{fig:transfer-finetune}
}
\caption{Loss curves related to LLM-gFTT transfer learning between Credit-g and Diabetes data domains. (a) Supervised finetuning of LLM using the source data set (Credit-g); (b) Finetuning of gFTT using the target data set (Diabetes) after cross-attention weight transfer. }
\label{fig:losses}
\vspace{-10pt}
\end{figure*}
\section {Experiments} \label{sec: experiments}
This section presents the experimental resources and steps used to evaluate the baseline and proposed models.
\subsection{Tabular data sets} 

We have identified nine source and five target publicly available tabular data sets from the OpenML repository~\citep{OpenML2013}. These data sets represent diverse application domains, including healthcare, finance, manufacturing, and software testing, to simulate realistic cross-domain scenarios. The selected tabular data sets have varying data types, a wide range of sample sizes ranging from 540 to 70,000, and feature dimensions between eight and 76. A summary of the 14 tabular data sets is presented in Table~\ref{tab:data_sets}. We enforce a strict disjoint feature space constraint to ensure that there are no common features between the source and target data sets. The strict condition differs from all previously known work on transfer learning, which considers common features between tabular data sets~\citep{Ye2024_CM2, Zhu2023_Xtab, nam2024_P2T, Wang2022_Transtab}. The source and target data sets are used to subsequently form ten pairs of source-target data with disjoint feature spaces.

\subsection{Baseline methods}\label{subsec: baselines}
We compare the performance of our proposed model against a diverse set of baselines, including traditional machine learning, state-of-the-art deep learning for tabular data, and transfer learning methods specifically designed for tabular data with and without LLM. Recent studies report the underperformance of LLMs in tabular data sets compared to traditional ML methods ~\citep{Dinh2022_LIFT, Yang2024_iTabLLM, Zhang2023_LLaMAGTL, Hegselmann2023_TabLLM, Han2024_featLLM,rabbani2025_transferLLM}. Unfortunately, traditional ML does not support transfer learning.  
Traditional machine learning (ML) includes logistic regression (LR), Multi-Layer Perceptron (MLP), and XGBoost. Deep learning baselines for tabular data without transfer learning include ResNet~\citep{Gorishniy2021_FTT} with residual connections adapted for tabular data, TabNet~\citep{Arik2021_TabNet} with an attention-based architecture for sequential feature selection, and FT-Transformer~\citep{Gorishniy2021_FTT} learning feature interactions through self-attention mechanisms. State-of-the-art transfer learning baselines include recent advances in cross-domain and multitable transfer learning. TransTab~\citep{Wang2022_Transtab} fine-tunes pretrained transformers for downstream tabular classification tasks. XTab~\citep{Zhu2023_Xtab} leverages multi-table pretraining to enhance generalization across multiple data tables. CM2~\citep{Ye2024_CM2} performs data representation transfer by aligning the statistical distributions between the source and target data sets. Additionally, we compare with Tabula-8B~\citep{Gardner2024_TABULA-8B}, an LLM pretrained on more than eight billion tabular rows and designed to support in-context learning for downstream tabular tasks without requiring task-specific fine-tuning. Furthermore, two baselines for LLM involving DistilGPT2 are presented in~\citep {rabbani2025_transferLLM}. One baseline is an LLM (single domain) that is fine-tuned on the target dataset only, and the other is an LLM (cross-domain) that is first fine-tuned on the source dataset and then adapted to the target. LLM methods involving in-context learning (ICL), such as P2T~\citep{nam2024_P2T}, require extensive manual effort to engineer prompts that capture the relationship between the source and target data at the individual sample level. In addition to this reason, we exclude ICL methods because of the unavailability of the source code.  


\subsection{Implementation and evaluation}

Transfer learning is a two-step process that involves pre-training and fine-tuning. Several baseline methods 
(CM2~\citep{Zhu2023_Xtab}, XTab~\citep{Ye2024_CM2}) provide a model pretrained on many tabular data sets, and we adopt these publicly available pretrained models and finetune them on each target data set. Otherwise, we pre-train a transfer learning model using our selected source data sets. Traditional baselines for machine learning and deep learning are evaluated directly using target data sets without any provision for transfer learning. The same training, validation, and test ratios are applied to the source and target data sets for pre-training and fine-tuning, respectively.

The training configuration of the DistilGPT2 model includes a learning rate of 3e-4, a weight decay of 0.01, a warm-up ratio of 0.1, and a batch size of 16. All other parameters follow the default settings in the Hugging Face implementation~\citep{huggingface_distilgpt2}. The fine-tuned DistilGPT2 with the lowest validation loss is selected for subsequent transfer learning. For transfer learning, a five-layer gFTT with weight transplants from the pretrained LLM (Figure~\ref{fig:lattle_arch}) is used for downstream target data classification. Each gFTT layer comprises eight attention heads and a feedforward network (FFN) with
a hidden dimension of 2048, ReLU activations, and a dropout layer. The hyperparameter distribution of the gFTT model used for our proposed transfer learning is tuned using Optuna~\citep{akiba2019optuna}. We obtain the area under the ROC curve (AUC) and classification accuracy (ACC) of the fine-tuned model using the test fold of the target data set. The mean AUC and mean ACC scores are obtained after repeating experiments for ten predefined random seeds.

\section{Results} \label{sec: results}

All experiments are conducted on a system running Ubuntu 20.04, equipped with an Intel(R) Core i9-13900F processor (32 cores, 5.60GHz), 64 GB of RAM, and an NVIDIA RTX 4090 GPU with 24 GB of memory. Deep learning models are implemented and trained using PyTorch, which natively supports parallel execution on multiple CPU cores to accelerate computation.

\begin{table*}[t]
\centering
\caption{Average area under the curve (AUC) scores with and without (no source pretraining) transfer learning. The best AUC scores of TABULA-8B are obtained from the corresponding paper~\citep{Gardner2024_TABULA-8B}. Cross-domain LLM fine-tunes a DistilGPT2 using source tabular data for downstream source-target transfer learning. \textit{* \small Indicates 32-shot (maximum possible) in-context learning AUC score using TABULA-8B.}}
\label{tab:transfer_learning}
\scalebox{0.48}{
\begin{tabular}{ccccccccccccccccccc}
\toprule
& \begin{tabular}[c]{@{}c@{}}Logistic\\ Regression\end{tabular} & XGBoost & MLP & ResNet & \begin{tabular}[c]{@{}c@{}}FT-\\ Transformer\end{tabular} & TabNet & \multicolumn{2}{c}{XTab} & \multicolumn{2}{c}{CM2} & \multicolumn{2}{c}{Tabula-8B} & \multicolumn{2}{c}{\begin{tabular}[c]{@{}c@{}}LLM\\ (Single-domain)\end{tabular}} & \multicolumn{2}{c}{TransTab} & \begin{tabular}[c]{@{}c@{}}LLM\\ (Cross-domain)\end{tabular} & \begin{tabular}[c]{@{}c@{}}LATTLE \\ (Proposed)\end{tabular} \\
\midrule
Target & AUC & AUC & AUC & AUC & AUC & AUC & Source & AUC & Source & AUC & Source & AUC & Source & AUC & Source & AUC & AUC & AUC \\
\midrule
\multirow{2}{*}{DB} 
& 0.822 (0.04) & 0.803 (0.03) & 0.798 (0.03) & 0.818 (0.03) & 0.788 (0.04) & 0.643 (0.03) & \multirow{10}{*}{\begin{tabular}[c]{@{}c@{}}52\\ AutoML\\ data\\ sets\end{tabular}} & 0.834 (0.02) & \multirow{10}{*}{\begin{tabular}[c]{@{}c@{}}Open\\ Tabs\\ 2000\\ data \\ sets\end{tabular}} & 0.803 (0.03) & \multirow{10}{*}{\begin{tabular}[c]{@{}c@{}}Eight\\ billion\\ tabular\\ rows\end{tabular}} & 0.680* & \multirow{10}{*}{\begin{tabular}[c]{@{}c@{}}Distil-\\ GPT2\end{tabular}} & 0.793 (0.04) & CD & 0.799 (0.04) & 0.653 (0.06) & 0.831 (0.04) \\
& 0.822 (0.04) & 0.803 (0.03) & 0.798 (0.03) & 0.818 (0.03) & 0.788 (0.04) & 0.643 (0.03) & & 0.834 (0.02) & & 0.803 (0.03) & & 0.680* & & 0.793 (0.04) & CG & 0.818 (0.04) & 0.639 (0.06) & 0.829 (0.04) \\
\multirow{2}{*}{VH} 
& 0.935 (0.01) & 0.925 (0.01) & 0.917 (0.01) & 0.855 (0.01) & 0.914 (0.01) & 0.794 (0.10) & & 0.935 (0.01) & & 0.893 (0.01) & & 0.484* & & 0.866 (0.02) & MF & 0.928 (0.01) & 0.711 (0.07) & 0.940 (0.00) \\
& 0.935 (0.01) & 0.925 (0.01) & 0.917 (0.01) & 0.855 (0.01) & 0.914 (0.01) & 0.794 (0.10) & & 0.935 (0.01) & & 0.893 (0.01) & & 0.484* & & 0.866 (0.02) & DG & 0.935 (0.01) & 0.633 (0.04) & 0.944 (0.00) \\
\multirow{2}{*}{CM} 
& 0.703 (0.02) & 0.726 (0.02) & 0.704 (0.02) & 0.695 (0.03) & 0.724 (0.03) & 0.681 (0.08) & & 0.721 (0.02) & & 0.729 (0.02) & & 0.352* & & 0.998 (0.00) & CH & 0.726 (0.01) & 0.959 (0.01) & 0.748 (0.02) \\
& 0.703 (0.02) & 0.726 (0.02) & 0.704 (0.02) & 0.695 (0.03) & 0.724 (0.03) & 0.681 (0.08) & & 0.721 (0.02) & & 0.729 (0.02) & & 0.352* & & 0.998 (0.00) & SK & 0.733 (0.02) & 0.954 (0.01) & 0.736 (0.03) \\
\multirow{2}{*}{PC1} 
& 0.826 (0.04) & 0.778 (0.04) & 0.825 (0.06) & 0.681 (0.04) & 0.810 (0.04) & 0.843 (0.08) & & 0.817 (0.05) & & 0.845 (0.10) & & 0.988* & & 0.996 (0.01) & CE & 0.825 (0.04) & 0.759 (0.11) & 0.879 (0.03) \\
& 0.826 (0.04) & 0.778 (0.04) & 0.825 (0.06) & 0.681 (0.04) & 0.810 (0.04) & 0.843 (0.08) & & 0.817 (0.05) & & 0.845 (0.10) & & 0.988* & & 0.996 (0.01) & SP & 0.812 (0.04) & 0.863 (0.02) & 0.886 (0.04) \\
\multirow{2}{*}{CB} 
& 0.816 (0.03) & 0.853 (0.04) & 0.603 (0.08) & 0.594 (0.07) & 0.677 (0.01) & 0.687 (0.05) & & 0.822 (0.03) & & 0.824 (0.05) & & -- & & 0.768 (0.04) & SP & 0.827 (0.03) & 0.686 (0.06) & 0.860 (0.03) \\
& 0.816 (0.03) & 0.853 (0.04) & 0.603 (0.08) & 0.594 (0.07) & 0.677 (0.01) & 0.687 (0.05) & & 0.822 (0.03) & & 0.824 (0.05) & & -- & & 0.768 (0.04) & SB & 0.825 (0.04) & 0.691 (0.07) & 0.865 (0.02) \\
\midrule
\begin{tabular}[c]{@{}c@{}}Avg.\\ Rank\end{tabular} 
& 5.4 (3.0) & 6.4 (3.0) & 8.3 (1.8) & 10.0 (3.3) & 8.9 (1.7) & 9.9 (2.8) & & 4.9 (3.3) & & 5.2 (1.7) & & 10.4 (4.5) & & 5.3 (3.8) & & 4.8 (2.5) & 8.6 (4.4) & 2.0 (0.9) \\
\midrule
Rank & 6 & 7 & 8 & 12 & 10 & 11 & & 3 & & 4 & & 13 & & 5 & & 2 & 9 & 1 \\
\bottomrule
\end{tabular}}
\end{table*}

\begin{table*}
\centering
\caption{Average classification accuracy (ACC) scores with and without (no source pretraining) transfer learning. The best ACC scores of Tabula-8B are not reported (N/R) in the corresponding paper~\citep{Gardner2024_TABULA-8B}. Cross-domain LLM finetunes a Distil-GPT using source tabular data for downstream source-target transfer learning.}
\label{tab:result-acc}
\scalebox{0.48}{
\begin{tabular}{ccccccccccccccccccc}
\toprule
& \begin{tabular}[c]{@{}c@{}}Logistic\\ Regression\end{tabular} & XGBoost & MLP & ResNet & \begin{tabular}[c]{@{}c@{}}FT-\\ Transformer\end{tabular} & TabNet & \multicolumn{2}{c}{XTab} & \multicolumn{2}{c}{CM2} & \multicolumn{2}{c}{Tabula-8B} & \multicolumn{2}{c}{\begin{tabular}[c]{@{}c@{}}LLM\\ (Single-domain)\end{tabular}} & \multicolumn{2}{c}{TransTab} & \begin{tabular}[c]{@{}c@{}}LLM\\ (Cross-domain)\end{tabular} & \begin{tabular}[c]{@{}c@{}}LATTLE \\ (Proposed)\end{tabular} \\
\midrule
Target & ACC & ACC & ACC & ACC & ACC & ACC & Source & ACC & Source & ACC & Source & ACC & Source & ACC & Source & ACC & ACC & ACC \\
\midrule
\multirow{2}{*}{DB}
& 0.747 (0.04) & 0.720 (0.04) & 0.753 (0.03) & 0.772 (0.03) & 0.754 (0.04) & 0.625 (0.05) & \multirow{10}{*}{\begin{tabular}[c]{@{}c@{}}52 \\ AutoML\\ data \\ sets\end{tabular}} & 0.764 (0.02) & \multirow{10}{*}{\begin{tabular}[c]{@{}c@{}}OpenTabs\\ 2000 \\ data\\ sets\end{tabular}} & 0.734 (0.02) & \multirow{10}{*}{\begin{tabular}[c]{@{}l@{}}Eight \\ billion\\ tabular\\ rows\end{tabular}} & N/R & \multirow{10}{*}{\begin{tabular}[c]{@{}l@{}}Distil-\\ GPT\end{tabular}} & 0.724 (0.03) & CD & 0.734 (0.04) & 0.610 (0.07) & 0.753 (0.03) \\
& 0.747 (0.04) & 0.720 (0.04) & 0.753 (0.03) & 0.772 (0.03) & 0.754 (0.04) & 0.625 (0.05) & & 0.764 (0.02) & & 0.734 (0.02) & & N/R & & 0.724 (0.03) & CG & 0.735 (0.04) & 0.603 (0.06) & 0.827 (0.04) \\
\multirow{2}{*}{VH}
& 0.736 (0.03) & 0.766 (0.03) & 0.754 (0.03) & 0.721 (0.04) & 0.767 (0.06) & 0.574 (0.01) & & 0.753 (0.02) & & 0.664 (0.02) & & N/R & & 0.630 (0.04) & MF & 0.776 (0.03) & 0.303 (0.09) & 0.942 (0.00) \\
& 0.736 (0.03) & 0.766 (0.03) & 0.754 (0.03) & 0.721 (0.04) & 0.767 (0.06) & 0.574 (0.01) & & 0.753 (0.02) & & 0.664 (0.02) & & N/R & & 0.630 (0.04) & DG & 0.766 (0.03) & 0.304 (0.10) & 0.940 (0.01) \\
\multirow{2}{*}{CM}
& 0.512 (0.04) & 0.507 (0.03) & 0.526 (0.02) & 0.536 (0.04) & 0.583 (0.06) & 0.766 (0.01) & & 0.653 (0.03) & & 0.552 (0.03) & & N/R & & 0.985 (0.01) & CH & 0.510 (0.02) & 0.864 (0.03) & 0.748 (0.02) \\
& 0.512 (0.04) & 0.507 (0.03) & 0.526 (0.02) & 0.536 (0.04) & 0.583 (0.06) & 0.766 (0.01) & & 0.653 (0.03) & & 0.552 (0.03) & & N/R & & 0.985 (0.01) & SK & 0.558 (0.03) & 0.840 (0.03) & 0.747 (0.02) \\
\multirow{2}{*}{PC1}
& 0.829 (0.04) & 0.829 (0.01) & 0.822 (0.01) & 0.630 (0.01) & 0.710 (0.04) & 0.743 (0.04) & & 0.717 (0.05) & & 0.765 (0.10) & & N/R & & 0.994 (0.01) & CE & 0.918 (0.04) & 0.930 (0.02) & 0.860 (0.03) \\
& 0.829 (0.04) & 0.829 (0.01) & 0.822 (0.01) & 0.630 (0.01) & 0.710 (0.04) & 0.743 (0.04) & & 0.717 (0.05) & & 0.765 (0.10) & & N/R & & 0.994 (0.01) & SP & 0.925 (0.01) & 0.928 (0.01) & 0.872 (0.04) \\
\multirow{2}{*}{CB}
& 0.746 (0.03) & 0.810 (0.04) & 0.673 (0.06) & 0.700 (0.05) & 0.557 (0.02) & 0.577 (0.04) & & 0.722 (0.03) & & 0.727 (0.05) & & N/R & & 0.718 (0.04) & SP & 0.778 (0.04) & 0.657 (0.05) & 0.853 (0.03) \\
& 0.746 (0.03) & 0.810 (0.04) & 0.673 (0.06) & 0.700 (0.05) & 0.557 (0.02) & 0.577 (0.04) & & 0.722 (0.03) & & 0.727 (0.05) & & N/R & & 0.718 (0.04) & SB & 0.764 (0.04) & 0.628 (0.09) & 0.846 (0.04) \\
\midrule
\begin{tabular}[c]{@{}c@{}}Avg.\\ Rank\end{tabular} 
& 6.6 (2.5) & 6.1 (4.1) & 7.3 (2.0) & 8.3 (2.9) & 7.4 (4.1) & 8.8 (3.4) & & 6.3 (2.3) & & 7.4 (1.5) & & N/R & & 5.2 (4.1) & & 4.7 (3.0) & 7.1 (4.9) & 2.3 (1.6) \\
\midrule
Rank & 6 & 4 & 8 & 11 & 9 & 12 & & 5 & & 9 & & N/R & & 3 & & 2 & 7 & 1 \\
\bottomrule
\end{tabular}}
\end{table*}

\begin{table*}[t]
\centering
\caption{Effects of different LLM to gFTT weight transplant strategies on source to target transfer learning AUC scores. $L$ is the uppermost and $0$ is the lowest attention layer.}
\label{tab:gftt-transfer-strategies}
\scalebox{0.75}{
\begin{tabular}{lccc}
\toprule
\textbf{Source $\rightarrow$ Target} &
\textbf{LLM}($W^L$) $\rightarrow$ \textbf{gFTT}($W^{0,1}$) & 
\textbf{LLM}($W^L$,$W^{L-1}$) $\rightarrow$ \textbf{gFTT}($W^{0,1}$) & 
\textbf{LLM}($W^L$) $\rightarrow$ \textbf{gFTT}($W^0$) [Proposed] \\
\midrule
CG$\rightarrow$ DB & 0.809 (0.05) & 0.804 (0.06) & {0.829 (0.04)} \\
DG$\rightarrow$ VH & 0.932 (0.00) & 0.936 (0.01) & {0.944 (0.00)} \\
SK$\rightarrow$ CM & 0.713 (0.03) & 0.719 (0.03) & {0.736 (0.03)} \\
\bottomrule
\end{tabular}
}
\end{table*}
\begin{figure*}[t]
\centerline{\includegraphics[width=0.65\textwidth]{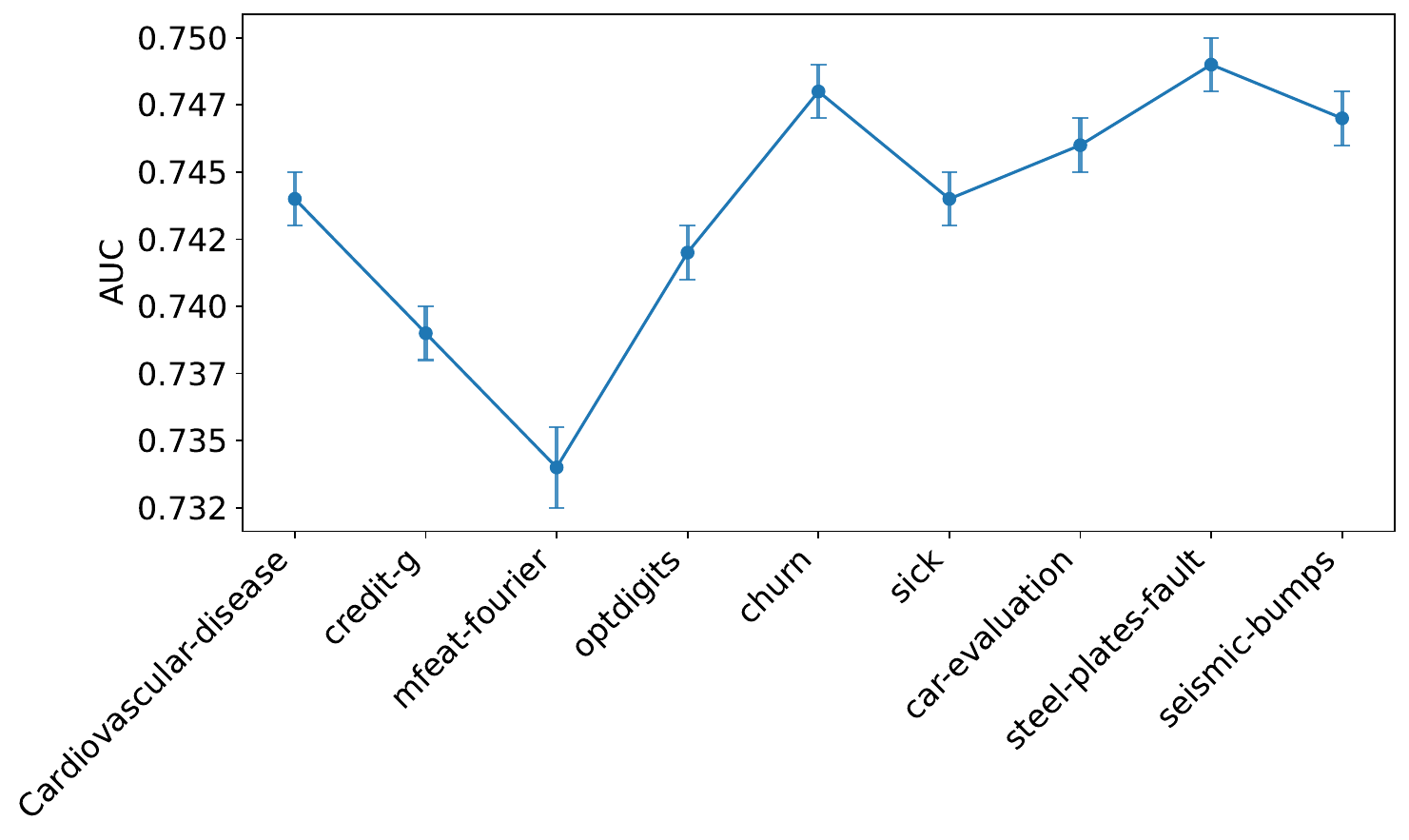}}
\caption{Effect of individual source data sets used in LLM pretraining on downstream transfer learning AUC scores. \textit{cmc} the target data set.}
\label{fig: auc-ablation}
\vspace{-10pt}
\end{figure*}

\subsection {Cross-domain data set pairs}

\begin{table}[h]
\centering
\caption{Hyperparameter search space for the proposed LATTLE method.}
\label{tab:hyperparameters-proposed}
\scalebox{0.85}{
\begin{tabular}{ll}
\toprule
\textbf{Parameter} & \textbf{Distribution} \\
\midrule
learning\_rate & LogUniform[1e$^{-5}$, 3e$^{-4}$] \\
batch\_size & Int[32, 128] (step=32) \\
weight\_decay & LogUniform[1e$^{-6}$, 1e$^{-2}$] \\
hidden\_dropout\_prob & Categorical\{0.0, 0.1, 0.2, 0.3, 0.4\} \\
warmup\_ratio & Categorical\{0.01, 0.05, 0.1\} \\
\bottomrule
\end{tabular}
}
\end{table}

\begin{table}[h]
\centering
\caption{Hyperparameter search space for traditional ML models.}
\label{tab:hyperparameters-ml}
\scalebox{0.8}{
\begin{tabular}{ll}
\toprule
\textbf{Parameter} & \textbf{Distribution} \\
\midrule
\multicolumn{2}{l}{\textit{Logistic Regression}} \\
\quad C & LogUniform[1e$^{-4}$, 1e$^{2}$] \\
\quad penalty & Categorical\{"l1", "l2"\} \\
\quad max\_iter & Int[100, 1000] \\
\midrule
\multicolumn{2}{l}{\textit{XGBoost}} \\
\quad max\_depth & Int[1, 10] \\
\quad learning\_rate & LogUniform[exp(-7), 1] \\
\quad n\_estimators & Int[100, 4000] \\
\bottomrule
\end{tabular}
}
\end{table}

\begin{table}[t]
\centering
\caption{Hyperparameter search space for deep learning baselines.}
\label{tab:hyperparameters-dl}
\scalebox{0.8}{
\begin{tabular}{ll}
\toprule
\textbf{Parameter} & \textbf{Distribution} \\
\midrule
\multicolumn{2}{l}{\textit{MLP, ResNet, FT-Transformer~\citep{Gorishniy2021_FTT}}} \\ \midrule
\quad learning\_rate & LogUniform[1e$^{-5}$, 3e$^{-4}$] \\ 
\quad batch\_size & Int[32, 128] (step=32) \\
\quad weight\_decay & LogUniform[1e$^{-6}$, 1e$^{-2}$] \\
\quad hidden\_dropout\_prob & Categorical\{0.0, 0.1, 0.2, 0.3, 0.4\} \\
\midrule
\multicolumn{2}{l}{\textit{TabNet~\citep{Arik2021_TabNet}}} \\ \midrule
\quad mask\_type & Categorical\{"entmax", "sparsemax"\} \\
\quad learning\_rate & LogUniform[1e$^{-3}$, 1e$^{-1}$] \\
\quad cat\_emb\_dim & Int[8, 32] (step=8) \\
\quad gamma & Uniform[1.0, 3.0] \\
\quad batch\_size & Categorical\{64, 128, 256\} \\
\midrule
\multicolumn{2}{l}{\textit{XTab~\citep{Zhu2023_Xtab}, TransTab~\citep{Wang2022_Transtab}, CM2~\citep{Ye2024_CM2}}} \\ \midrule
\quad learning\_rate & LogUniform[1e$^{-5}$, 3e$^{-4}$] \\
\quad batch\_size & Int[32, 128] (step=32) \\
\quad weight\_decay & LogUniform[1e$^{-6}$, 1e$^{-2}$] \\
\quad hidden\_dropout\_prob & Categorical\{0.0, 0.1, 0.2, 0.3, 0.4\} \\
\quad warmup\_ratio & Categorical\{0.01, 0.05, 0.1\} \\
\bottomrule
\end{tabular}
\vspace{-10pt}
}
\end{table}

We create ten data set pairs using source and target examples from Table~\ref{tab:data_sets}. The source and target data sets are of different domains are paired after ensuring no common features in between. Following the abbreviations in Table~\ref{tab:data_sets}, the ten pairs of cross-domain data sets include CG-DB (banking-metabolic disease), CD-DB (heart disease-metabolic disease), MF-VH (shape measurements-automotive), DG-VH (optical digits-automotive), CH-CM (telecommunication-demographics), SK-CM (thyroid disease-demographics), SP-PC1 (manufacturing-software testing), CE-PC1 (car pricing-software testing), SP-CB (manufacturing-industrial design), and SB-CB (hazard monitoring-industrial design).




\subsection{Model training and selection}

All experiments use a consistent data split strategy for both source and target data sets, allocating 70\% of each data set for training, 10\% for validation, and 20\% for testing. Using predefined random seeds, these data folds are sampled ten times following the 70-10-20 data split strategy. First, the source training data fold is used to fine-tune an open-source LLM (DistilGPT2) with default parameters for 200 epochs. Figure \ref{fig:sup-pretrain} shows the model convergence while fine-tuning the LLM using the source tabular data. In downstream transfer learning, we have tuned the gFTT hyperparameters using 100 Optuna trials and the target data set. The validation set is used to select the gFTT model with the lowest validation loss during 150 epochs of training. The distribution of the tuned hyperparameters of our proposed model is presented in Table~\ref{tab:hyperparameters-proposed}

The selected model is used to report the transfer learning performance using the test fold of the target data set. Figure ~\ref{fig:transfer-finetune} shows that the gFTT training and validation loss curves converge after 80 epochs. The search space for the tuned hyperparameters of traditional machine learning and deep learning baselines is shown in Tables~\ref{tab:hyperparameters-ml} and \ref{tab:hyperparameters-dl}, respectively.

\subsection{Transfer learning performance}
The performance of the proposed transfer learning based on cross-attention is compared with 12 traditional ML and DL baseline methods, including the state-of-the-art transfer learning methods (SOTA) for tabular data with and without LLM (Section~\ref{subsec: baselines}). Table~\ref{tab:transfer_learning} shows that traditional ML methods (LR, XGBoost) outperform recent DL methods (FT-Transformer, TabNet) proposed for tabular data based on average performance rank. Recently proposed transfer learning methods for tabular data, including XTab (rank: 3), CM2 (rank: 4), and TransTab (rank: 2), have all performed better than the ML and DL methods for tabular data. Among the transfer learning methods, TransTab outperforms the other two (XTab and CM2). In particular, TransTab requires a single source data set, whereas XTab and CM2 are pretrained on 52 and 2000 tabular data sets, respectively. 

In contrast, LLM-based transfer learning methods perform poorly compared to the transfer learning methods proposed for tabular data. The Tabula-8B model, pretrained on eight billion rows, demonstrates one of the worst classification performances on target data sets (rank:13). For the VH data set, Tabula-8B performs worse than random guessing, while all other competitive baselines achieve AUC scores above 0.90. The best classification scores in the TABULA-8B paper are obtained and reported by prompting the pretrained model with 32 shot target samples~\citep{Gardner2024_TABULA-8B}. Another baseline shows that end-to-end fine-tuning an LLM (DistilGPT2) using cross-domain source–target pairs also underperforms (rank: 9) compared to ML and transfer learning methods without LLM. In contrast, end-to-end fine-tuning of DistilGPT2 using a single domain target tabular data results in better classification performance than the ML and DL methods. This result suggests that LLMs are not optimally designed for transfer learning of tabular data. The comparison between Tabula-8B and DistilGPT2 suggests that downstream fine-tuning using target data is preferred over ICL, that is, learning with a few shots using text prompts. Our proposed LATTLE method, on average, is the best ranked (average rank: 2.0 (0.9)) compared to the second-best TransTab method (average rank: 4.8 (2.5)). Although TransTab is a gFTT-based method, our proposed method effectively uses LLM weight transplants to improve the performance of gFTT.

The performance of the proposed and baseline methods in terms of classification accuracy is mostly consistent with that reported using AUC scores. The proposed LATTLE method remains the best method (average rank: 2.3 (1.6)), followed by TransTab (average rank: 4.7 (3.0)). However, XGBoost outperforms the XTab and CM2 methods in terms of ACC in contrast to the rank order obtained using AUC. Likewise, LLM (single domain) ranks better than XTab and CM2 unlike the performance rank obtained using AUC. In this context, it can be argued that methods that are susceptible to data imbalance are likely to report inflated performance when accuracy is used as the performance metric.






\subsection{Ablation study}

The selection of attention layers for weight transfer may affect transfer learning performance. Table~\ref{tab:gftt-transfer-strategies} shows the effect of using LLM weights from the top ($W^L$) or two top ($W^L$, $W^{L-1}$) attention layers in the lowest attention layers ($W^0$, $W^1$) of gFTT. The differences in the mean AUC scores are minimal, whereas the proposed method ({LLM}($W^L)$$\rightarrow$ gFTT($W^0$)) yields the best performance. Figure~\ref{fig: auc-ablation} shows the effect of different source data sets on fine-tuning the LLM for transfer learning. The downstream AUC scores remain mostly stable for varying source data sets where the AUC scores are within 1.5\%.

\section{Discussion of the results} \label{sec: Discussion}
This work presents one of the first cross-domain transfer learning between tabular data sets with completely disjoint feature sets. The key findings of this paper are as follows. First, the proposed LATTLE method demonstrates that effective transfer learning through an LLM can achieve better performance than state-of-the-art ML and DL methods for tabular data. Second, weight transplants from a selective attention layer of an LLM are more effective for cross-domain attention learning than end-to-end fine-tuning of an LLM. Third, downstream transfer learning of tabular data should be performed using a tabular transformer model (e.g., gFTT) instead of finetuning a language model developed for text data. Fourth, we find that a single source data set is sufficient for tabular transfer learning when cross-attention is used through weight transplants. The proposed cross-table attention learning alleviates the need for large-scale pertaining using tens (e.g., XTab~\citep{Zhu2023_Xtab}) or thousands (e.g., CM2~\citep{Ye2024_CM2}) of tables. Fifth, the comparison of baselines indicates that fine-tuning on the target data set is superior to popular in-context learning via prompt engineering.

It is well-established in the literature that effective transfer learning requires a model pretrained with large data sets. The requirement of large data samples is underscored in the development of large vision and large language models. The same hypothesis is adopted in recent studies to achieve transfer learning of tabular data, which involves building large pretrained models~\citep {Gardner2024_TABULA-8B, Zhu2023_Xtab, Ye2024_CM2}. In particular, large pretrained models for text and image domains have been effective because knowledge synthesis from multiple data sources can seamlessly occur through shared patterns and semantics in homogeneous image or text feature spaces. 

However, the inductive bias for tabular data is different from that of image and text data{~\citep{Katzir2021, Borisov2022_Survey}. The same pretraining strategy using large tabular data samples from disparate sources with heterogeneous feature spaces may not seamlessly integrate generalizable knowledge into one model, leading to confusion and hallucination in downstream fine-tuning for transfer learning.  Our proposed weight transplant method suggests that context learning, by attending to different domains, prioritizing the inductive bias, and learning structural information of tabular data are preferred approaches to transfer learning rather than fine-tuning a large pre-trained model.

The proposed LATTLE algorithm learns the relationship between tabular data sets for transfer learning in three ways. First, fine-tuning the LLM coupled with gFTT aligns tabular data representations with language representations. This text-to-tabular data alignment prepares LLM for transfer learning of tabular data. Second, the attention between LLM and gFTT through weight transplant is used to learn the contextual relationship between source and target data, which improves the downstream learning of the target tabular data. Instead of mining transferable knowledge from a large pretrained model, the proposed method leverages context learning and general representation of heterogeneous data types. Third, downstream fine-tuning is achieved using a tabular transformer model with the necessary inductive bias, which may not be attainable using language models.  

The observation in this paper confirms the finding of a recent study~\citep{spinaci2025ConTextTab} about the limitation of ICL due to token inefficiency, limited scalability, and subjectivity in prompt engineering. Prompt engineering and ICL are used because of the convenience of probing powerful and up-to-date LLMs without the computational requirement of fine-tuning these large models. However, transforming structured data tables into documents involves manual efforts and loses important structural information about the data. Generative models such as LLMs are developed to output text responses, which would intuitively require customization to adapt to tabular data classification tasks. In this context, the limitation of ICL can be overcome by directly fine-tuning the weights of an LLM to perform a target learning task. However, knowledge transfer and alignment of the representation between text and tabular data are open problems.

Our $query$ representation of target data in the $key$ and $value$ representations of a fine-tuned LLM establishes a useful contextual bridge between language and tabular data representations. The robustness of the algorithm to varying source data sets suggests that our method does not rely on the domain knowledge of the source data. In cross-domain transfer learning, depending solely on the source data is impractical when the target data is of a different domain without any shared features. Instead, the context between domains is learned through cross-attention, using a distilled version of basic LLM. A recent study\citep{wydmuch2024_LLM-RDB} has concluded that focus on context is far more important for tabular data than using larger LLMs with higher parameter counts.

The selective finetuning and transplant of attention layers of an LLM suggest a potential solution to addressing the model's low performance due to hallucination. An end-to-end LLM finetuning using tabular data and a classifier head may disrupt the pretrained knowledge from a large language corpus using an autoregressive loss. This hypothesis explains why end-to-end fine-tuning of the same LLM has ranked ninth in Table~\ref{tab:transfer_learning}. Inspired by the hierarchy of knowledge distribution in a deep neural network, the lower attention layers are frozen to retain the model's general knowledge from the language corpus while fine-tuning the upper layer to learn and adapt to new knowledge from the source data{~\citep{houlsby2019_upperlayerVSlower}.

In this context, the ablation study shows that the uppermost attention layer of an LLM is sufficient for transfer learning. Hierarchically, the uppermost attention layer of the LLM contains the most task-specific knowledge, especially due to its close proximity to the tabular transformer model (gFTT) in the LLM-gFTT framework. The insensitivity of the transfer learning performance to the source data set suggests that the projection weights of $key$ and $value$ effectively learn the context between LLM and gFTT. 
Therefore, our proposed method circumvents the need for models pretrained using large data samples, which could potentially lead to data memorization and hallucination. 

\subsection{Limitations}
Despite the best performance rank, the proposed LATTLE algorithm underperforms in several data sets due to unexplained data-specific characteristics. Although recent LLMs can memorize public data sets, the possibility is minimal with a distilled version of one of the earliest GPTs. The computational and memory requirements for training LLMs are considerably high, which limited us to using the distilled version of a GPT. More recent and larger language models may yield better performance than what we report. However, we cannot eliminate the possibility of data memorization and hallucination of LLM after end-to-end training. Despite heavy computation and large data provision, current large pretrained models have failed to achieve satisfactory performance. Therefore, the scaling of the proposed model remains an open problem.

\section {Conclusions} \label{sec: Conclusions}
This article presents a novel LLM-based cross-attention method to learn relationships between tabular data sets. The results reveal that the proposed cross-attention method makes a lightweight LLM sufficient to outperform the state-of-the-art deep and transfer learning methods proposed for tabular data. Experimental results demonstrate the effectiveness of one of the first frameworks proposed for learning relationships between tabular data of two different domains without requiring shared features. Our findings indicate that a single source data set is sufficient for pretraining when an LLM is used, thereby eliminating the need for large-scale tabular data sets.

\section{Acknowledgements} \label{sec: Acknowledgements}

The research reported in this publication was partially supported by the US National Science Foundation (NSF) award \# 2431058 and received support from the Air Force Office of Scientific Research under Grant Number W911NF-23-1-0170. The content is solely the responsibility of the authors and should not be interpreted as representing the official policies, either expressed or implied, of the Army Research Office or the U.S. Government.

\bibliographystyle{elsarticle-num}
\bibliography{LLM, Tabular}

@inproceedings{houlsby2019_upperlayerVSlower,
  title={Parameter-efficient transfer learning for NLP},
  author={Houlsby, Neil and Giurgiu, Andrei and Jastrzebski, Stanislaw and Morrone, Bruna and De Laroussilhe, Quentin and Gesmundo, Andrea and Attariyan, Mona and Gelly, Sylvain},
  booktitle={International conference on machine learning},
  pages={2790--2799},
  year={2019},
  organization={PMLR}
}

@article{hwang2025_freezinglower,
  title={Exploring Selective Layer Freezing Strategies in Transformer Fine-Tuning: NLI Classifiers with Sub-3B Parameter Models},
  author={Hwang, Taewook and Seo, Hyein and Jung, Jeesu and Jung, Sangkeun},
  journal={Applied Sciences},
  volume={15},
  number={19},
  pages={10434},
  year={2025},
  publisher={MDPI}
}

@inproceedings{dai2022,
    title = "Knowledge Neurons in Pretrained Transformers",
    author = "Dai, Damai  and
      Dong, Li  and
      Hao, Yaru  and
      Sui, Zhifang  and
      Chang, Baobao  and
      Wei, Furu",
    editor = "Muresan, Smaranda  and
      Nakov, Preslav  and
      Villavicencio, Aline",
    booktitle = "Proceedings of the 60th Annual Meeting of the Association for Computational Linguistics (Volume 1: Long Papers)",
    month = may,
    year = "2022",
    address = "Dublin, Ireland",
    publisher = "Association for Computational Linguistics",
    url = "https://aclanthology.org/2022.acl-long.581/",
    doi = "10.18653/v1/2022.acl-long.581",
    pages = "8493--8502"
}

@article{wydmuch2024_LLM-RDB,
  title={Tackling prediction tasks in relational databases with LLMs},
  author={Wydmuch, Marek and Borchmann, {\L}ukasz and Grali{\'n}ski, Filip},
  journal={arXiv preprint arXiv:2411.11829},
  year={2024}
}

@article{chen2022_limitation-ICL,
  title={Large language models are few (1)-shot table reasoners},
  author={Chen, Wenhu},
  journal={arXiv preprint arXiv:2210.06710},
  year={2022}
}

@misc{huggingface_distilgpt2,
  author = {{Hugging Face}},
  title = {DistilGPT2},
  year = {2019},
  howpublished = {\url{https://huggingface.co/distilgpt2}},
  note = {Accessed: 2024-09-20}
}

@article{radford2019_GPT2,
  title={Language Models are Unsupervised Multitask Learners},
  author={Radford, Alec and Wu, Jeff and Child, Rewon and Luan, David and Amodei, Dario and Sutskever, Ilya},
  year={2019}
}

@inproceedings{Huertas2024ImproveLGMonFewShots,
    author={Carlos Huertas},
    title={Gradient Boosting Trees and Large Language Models for Tabular Data Few-Shot Learning},
    booktitle={19th Conference on Computer Science and Intelligence Systems},
    year={2024},
    publisher={IEEE},
    series={Annals of Computer Science and Information Systems}
}

@article{brown2020_language-models-fews-hot-learners,
  title={Language models are few-shot learners},
  author={Brown, Tom and Mann, Benjamin and Ryder, Nick and Subbiah, Melanie and Kaplan, Jared D and Dhariwal, Prafulla and Neelakantan, Arvind and Shyam, Pranav and Sastry, Girish and Askell, Amanda and others},
  journal={Advances in neural information processing systems},
  volume={33},
  pages={1877--1901},
  year={2020}
}

@INPROCEEDINGS{Abdul2024_TabLM,
  author={Shakir, Wajiha Abdul},
  booktitle={2024 1st International Conference on Emerging Technologies for Dependable Internet of Things (ICETI)}, 
  title={Benchmarking TabLM: Evaluating the Performance of Language Models Against Traditional Machine Learning in Structured Data Tasks}, 
  year={2024},
  volume={},
  number={},
  pages={1-10},
  doi={10.1109/ICETI63946.2024.10777155}}

@inproceedings{
zeng2025llm-same-domain-TL,
title={{LLM} Embeddings Improve Test-time Adaptation to Tabular \$Y|X\$-Shifts},
author={Yibo Zeng and Jiashuo Liu and Henry Lam and Hongseok Namkoong},
booktitle={NeurIPS 2024 Third Table Representation Learning Workshop},
year={2024},
url={https://openreview.net/forum?id=DQn3ubkN2B}
}

@article{shi2025_LATTE,
  title={Latte: Transfering LLMsLatent-level Knowledge for Few-shot Tabular Learning},
  author={Shi, Ruxue and Gu, Hengrui and Ye, Hangting and Dai, Yiwei and Shen, Xu and Wang, Xin},
  journal={arXiv preprint arXiv:2505.05237},
  year={2025}
}

@inproceedings{rabbani2025_transferLLM,
   author = {Shourav B. Rabbani and Ibna Kowsar and Manar D. Samad},
   doi = {10.1109/ICECE64886.2024.11024938},
   isbn = {979-8-3315-3178-2},
   booktitle = {2024 13th International Conference on Electrical and Computer Engineering (ICECE)},
   month = {12},
   pages = {287-292},
   publisher = {IEEE},
   title = {Transfer Learning of Tabular Data by Finetuning Large Language Models},
   url = {https://ieeexplore.ieee.org/document/11024938/},
   year = {2024}
}

@inproceedings{nam2024_P2T,
archivePrefix = {arXiv},
arxivId = {2408.11063},
author = {Nam, Jaehyun and Song, Woomin and Park, Seong Hyeon and Tack, Jihoon and Yun, Sukmin and Kim, Jaehyung and Oh, Kyu Hwan and Shin, Jinwoo},
booktitle = {First Conference on Language Modeling},
eprint = {2408.11063},
month = {aug},
title = {{Tabular Transfer Learning via Prompting LLMs}},
url = {https://openreview.net/forum?id=bo4pauxnIR},
year = {2024}
}

@article{Yang2024_iTabLLM,
journal = {arXiv preprint arXiv:2403.20208},
author = {Yang, Yazheng and Wang, Yuqi and Sen, Sankalok and Li, Lei and Liu, Qi},
eprint = {2403.20208},
title = {{Unleashing the Potential of Large Language Models for Predictive Tabular Tasks in Data Science}},
url = {http://arxiv.org/abs/2403.20208},
year = {2024}
}

@inproceedings{Wang2023a_MediTab,
author = {Wang, Zifeng and Gao, Chufan and Xiao, Cao and Sun, Jimeng},
title = {MediTab: scaling medical tabular data predictors via data consolidation, enrichment, and refinement},
year = {2024},
isbn = {978-1-956792-04-1},
url = {https://doi.org/10.24963/ijcai.2024/670},
doi = {10.24963/ijcai.2024/670},
booktitle = {Proceedings of the Thirty-Third International Joint Conference on Artificial Intelligence},
articleno = {670},
numpages = {9},
location = {Jeju, Korea},
series = {IJCAI '24}
}

@article{Zhang2023_LLaMAGTL,
journal = {arXiv preprint arXiv:2310.07338},
author = {Zhang, Han and Wen, Xumeng and Zheng, Shun and Xu, Wei and Bian, Jiang},
eprint = {2310.07338},
month = {oct},
pages = {1--22},
title = {{Towards Foundation Models for Learning on Tabular Data}},
url = {http://arxiv.org/abs/2310.07338},
year = {2023}
}

@article{Hegselmann2023_TabLLM,
archivePrefix = {arXiv},
arxivId = {2210.10723},
author = {Hegselmann, Stefan and Buendia, Alejandro and Lang, Hunter and Agrawal, Monica and Jiang, Xiaoyi and Sontag, David},
eprint = {2210.10723},
issn = {26403498},
journal = {Proceedings of Machine Learning Research},
pages = {5549--5581},
title = {{TabLLM: Few-shot Classification of Tabular Data with Large Language Models}},
volume = {206},
year = {2023}
}

@article{Gardner2024_TABULA-8B,
archivePrefix = {arXiv},
arxivId = {2406.12031},
author = {Gardner, Joshua P and Perdomo, Juan Carlos and Schmidt, Ludwig},
eprint = {2406.12031},
journal = {The Thirty-eighth Annual Conference on Neural Information Processing Systems},
pages = {1--38},
title = {{Large Scale Transfer Learning for Tabular Data via Language Modeling}},
url = {https://openreview.net/forum?id=WH5blx5tZ1},
year = {2024}
}

@article{Dinh2022_LIFT,
archivePrefix = {arXiv},
arxivId = {2206.06565},
author = {Dinh, Tuan and Zeng, Yuchen and Zhang, Ruisu and Lin, Ziqian and Gira, Michael and Rajput, Shashank and Sohn, Jy Yong and Papailiopoulos, Dimitris and Lee, Kangwook},
eprint = {2206.06565},
isbn = {9781713871088},
issn = {10495258},
journal = {Advances in Neural Information Processing Systems},
number = {NeurIPS},
title = {{LIFT: Language-Interfaced Fine-Tuning for Non-Language Machine Learning Tasks}},
volume = {35},
year = {2022}
}

@article{Han2024_featLLM,
journal = {arXiv preprint arXiv:2404.09491},
author = {Han, Sungwon and Yoon, Jinsung and Arik, Sercan O and Pfister, Tomas},
eprint = {2404.09491},
month = {apr},
title = {{Large Language Models Can Automatically Engineer Features for Few-Shot Tabular Learning}},
url = {http://arxiv.org/abs/2404.09491},
year = {2024}
}

@article{spinaci2025ConTextTab,
  title={ConTextTab: A Semantics-Aware Tabular In-Context Learner},
  author={Spinaci, Marco and Polewczyk, Marek and Schambach, Maximilian and Thelin, Sam},
  journal={arXiv preprint arXiv:2506.10707},
  year={2025}
}

@inproceedings{akiba2019optuna,
  title={{O}ptuna: A Next-Generation Hyperparameter Optimization Framework},
  author={Akiba, Takuya and Sano, Shotaro and Yanase, Toshihiko and Ohta, Takeru and Koyama, Masanori},
  booktitle={The 25th ACM SIGKDD International Conference on Knowledge Discovery \& Data Mining},
  pages={2623--2631},
  year={2019}
}

@article{OpenML2013,
      author = {Joaquin Vanschoren and Jan N. van Rijn and Bernd Bischl and Luis Torgo},
      title = {OpenML: networked science in machine learning},
      journal = {SIGKDD Explorations},
      volume = {15},
      number = {2},
      year = {2013},
      pages = {49-60},
      url = {},
      doi = {10.1145/2641190.2641198},
      publisher = {ACM}
    }

@article{rabbani2024_gceals,
  title={Deep clustering of tabular data by weighted Gaussian distribution learning},
  author={Rabbani, Shourav B and Medri, Ivan V and Samad, Manar D},
  journal={Neurocomputing},
  volume={623},
  pages={129359},
  year={2025},
  publisher={Elsevier}
}

@inproceedings{kowsar2023buet_deepCluster,
  title={Deep Clustering of Electronic Health Records Tabular Data for Clinical Interpretation},
  author={Kowsar, Ibna and Rabbani, Shourav B and Akhter, Kazi Fuad B and Samad, Manar D},
  booktitle={2023 IEEE International Conference on Telecommunications and Photonics (ICTP)},
  pages={01--05},
  year={2023},
  organization={IEEE}
}

@article{ke2017lightgbm,
  title={Lightgbm: A highly efficient gradient boosting decision tree},
  author={Ke, Guolin and Meng, Qi and Finley, Thomas and Wang, Taifeng and Chen, Wei and Ma, Weidong and Ye, Qiwei and Liu, Tie-Yan},
  journal={Advances in neural information processing systems},
  volume={30},
  year={2017}
}

@inproceedings{Ye2024_CM2,
author = {Ye, Chao and Lu, Guoshan and Wang, Haobo and Li, Liyao and Wu, Sai and Chen, Gang and Zhao, Junbo},
title = {Towards Cross-Table Masked Pretraining for Web Data Mining},
year = {2024},
isbn = {9798400701719},
publisher = {Association for Computing Machinery},
address = {New York, NY, USA},
url = {https://doi.org/10.1145/3589334.3645707},
doi = {10.1145/3589334.3645707},
booktitle = {Proceedings of the ACM Web Conference 2024},
pages = {4449–4459},
numpages = {11},
location = {Singapore, Singapore},
series = {WWW '24}
}

@inproceedings{
levin2023_Tabular-transfer,
title={Transfer Learning with Deep Tabular Models},
author={Roman Levin and Valeriia Cherepanova and Avi Schwarzschild and Arpit Bansal and C. Bayan Bruss and Tom Goldstein and Andrew Gordon Wilson and Micah Goldblum},
booktitle={The Eleventh International Conference on Learning Representations },
year={2023},
url={https://openreview.net/forum?id=b0RuGUYo8pA}
}

@inproceedings{
hollmann2022_TabPFN,
title={Tab{PFN}: A Transformer That Solves Small Tabular Classification Problems in a Second},
author={Noah Hollmann and Samuel M{\"u}ller and Katharina Eggensperger and Frank Hutter},
booktitle={The Eleventh International Conference on Learning Representations },
year={2023},
url={https://openreview.net/forum?id=cp5PvcI6w8_}
}

@article{Borisov2022_Survey,
  title={Deep neural networks and tabular data: A survey},
  author={Borisov, Vadim and Leemann, Tobias and Se{\ss}ler, Kathrin and Haug, Johannes and Pawelczyk, Martin and Kasneci, Gjergji},
  journal={IEEE Transactions on Neural Networks and Learning Systems},
  year={2022},
  publisher={IEEE}
}

@article{Katzir2021,
author = {Katzir, Liran and Elidan, Gal and El-Yaniv, Ran},
isbn = {9781606924358},
journal = {International Conference on Learning Representations (ICLR)},
number = {December},
pages = {1066--1071},
title = {{Net-DNF: Effective deep modeling of tabular data}},
volume = {125},
year = {2021}
}

@article{Gorishniy2021_FTT,
archivePrefix = {arXiv},
arxivId = {2106.11959},
author = {Gorishniy, Yury and Rubachev, Ivan and Khrulkov, Valentin and Babenko, Artem},
eprint = {2106.11959},
isbn = {9781713845393},
issn = {10495258},
journal = {Advances in Neural Information Processing Systems},
month = {dec},
number = {NeurIPS},
pages = {18932--18943},
title = {{Revisiting Deep Learning Models for Tabular Data}},
url = {https://github.com/yandex-research/rtdl. http://arxiv.org/abs/2106.11959},
volume = {23},
year = {2021}
}

@inproceedings{
Wang2022_Transtab,
title={TransTab: Learning Transferable Tabular Transformers Across Tables},
author={Zifeng Wang and Jimeng Sun},
booktitle={Advances in Neural Information Processing Systems},
editor={Alice H. Oh and Alekh Agarwal and Danielle Belgrave and Kyunghyun Cho},
year={2022},
url={https://openreview.net/forum?id=A1yGs_SWiIi}
}

@InProceedings{Zhu2023_Xtab,
  title = 	 {{XT}ab: Cross-table Pretraining for Tabular Transformers},
  author =       {Zhu, Bingzhao and Shi, Xingjian and Erickson, Nick and Li, Mu and Karypis, George and Shoaran, Mahsa},
  booktitle = 	 {Proceedings of the 40th International Conference on Machine Learning},
  pages = 	 {43181--43204},
  year = 	 {2023},
  editor = 	 {Krause, Andreas and Brunskill, Emma and Cho, Kyunghyun and Engelhardt, Barbara and Sabato, Sivan and Scarlett, Jonathan},
  volume = 	 {202},
  series = 	 {Proceedings of Machine Learning Research},
  month = 	 {23--29 Jul},
  publisher =    {PMLR},
  url = 	 {https://proceedings.mlr.press/v202/zhu23k.html}
}

@inproceedings{Arik2021_TabNet,
archivePrefix = {arXiv},
arxivId = {1908.07442},
author = {Arik, Sercan O. and Pfister, Tomas},
booktitle = {Proceedings of the AAAI Conference on Artificial Intelligence},
eprint = {1908.07442},
month = {aug},
pages = {6679--6687},
title = {{TabNet: Attentive Interpretable Tabular Learning}},
url = {http://arxiv.org/abs/1908.07442},
year = {2021}
}

@inproceedings{chen2016xgboost,
address = {New York, NY, USA},
author = {Chen, Tianqi and Guestrin, Carlos},
booktitle = {Proceedings of the 22nd ACM SIGKDD International Conference on Knowledge Discovery and Data Mining},
doi = {10.1145/2939672.2939785},
isbn = {9781450342322},
month = {aug},
pages = {785--794},
publisher = {ACM},
title = {{XGBoost}},
url = {https://dl.acm.org/doi/10.1145/2939672.2939785},
volume = {13-17-Augu},
year = {2016}
}

@article{Grinsztajn2022,
  title={Why do tree-based models still outperform deep learning on typical tabular data?},
  author={Grinsztajn, L{\'e}o and Oyallon, Edouard and Varoquaux, Ga{\"e}l},
  journal={Advances in neural information processing systems},
  volume={35},
  pages={507--520},
  year={2022}
}

\end{document}